\pdfoutput=1
\documentclass[sigconf,nonacm]{acmart}

\usepackage{adjustbox}
\usepackage{graphicx}
\usepackage{tabularx}
\usepackage{amsmath}   
\usepackage{amssymb}   
\usepackage{array}     
\usepackage{booktabs}
\usepackage{multirow}
\usepackage{enumitem}
\usepackage{subcaption}
\usepackage{pifont}    
\usepackage{tikz}
\usepackage{placeins}

\newcommand{\xmark}{\ding{55}} 

\AtBeginDocument{%
  }

\setcopyright{none}
\renewcommand\footnotetextcopyrightpermission[1]{}
\begin{document}

\title[EVAS]{EVAS: Efficient Multimodal Temporal Forgery Localization via Audio-Visual Synergy and Steered Boundary Calibration}



\author{Shen Shen}
\authornote{Both authors contributed equally to this research.}
\email{sshen@stu.suda.edu.cn}
\affiliation{%
  \institution{Soochow University}
  \city{Suzhou}
  \country{China}
}

\author{Quan Zhang}
\authornotemark[1]
\email{zhangqua22@tsinghua.org.cn}
\affiliation{%
  \institution{Tsinghua University}
  \city{Beijing}
  \country{China}
}

\author{Dan Jiang}
\affiliation{%
  \institution{Tsinghua University}
  \city{Beijing}
  \country{China}
}

\author{Ke Zhang}
\authornote{Corresponding author.}
\email{kzhang19@suda.edu.cn}
\affiliation{%
  \institution{Soochow University}
  \city{Suzhou}
  \country{China}
}

\renewcommand{\shortauthors}{Shen et al.}

\begin{abstract}
The rapid proliferation of artificial intelligence-generated content necessitates reliable multimodal forensics. Beyond video-level binary classification, precisely localizing sparsely distributed forged segments in long-form videos remains a critical challenge. This task is particularly difficult when manipulations are subtly embedded and cross-modal signals are weak and temporally diffuse. To address these challenges, we propose EVAS, an end-to-end multimodal framework for temporal forgery localization. At its core, a Multi-Stage Audio-Visual Synergy mechanism facilitates progressive cross-modal interaction to learn deep multimodal forensic representations and capture high-order semantic traces of sparse manipulations. Furthermore, we introduce a Boundary-Aware Refinement strategy to achieve steered boundary calibration. By incorporating invalid-frame masking, this strategy suppresses ambiguous regions and sharpens transition predictions. We adopt a decoupled training paradigm with auxiliary heads to disentangle representation learning from inference objectives, enhancing model generalization and stability. Additionally, a lightweight HourglassFFN is incorporated to reduce computational overhead. Extensive experiments demonstrate that EVAS achieves state-of-the-art average localization accuracy and average recall across three benchmark datasets, validating its effectiveness for fine-grained temporal forgery localization.
\end{abstract}




\maketitle
\pagestyle{plain}
\section{Introduction}
The rapid advancement of Artificial Intelligence Generated Content (AIGC) \cite{ref1} has significantly lowered the barrier for creating hyper-realistic deepfakes, posing a severe threat to information integrity \cite{ref2,ref3}. Consequently, multimedia forensics is shifting from video-level binary classification\cite{ref9, ref10, ref11, ref12, ref13, ref14, ref15, ref16, ref17, ref18} to the more granular task of Temporal Forgery Localization (TFL) \cite{ref6}, which increasingly relies on multimodal cues to detect subtle inconsistencies. This shift addresses temporal sparse attacks\cite{ref46, ref48} where adversaries inject brief and concealed forged segments into legitimate footage. Such sparse manipulations easily evade global statistical detection systems, threatening real-world verification processes\cite{ref22, ref52}.

Despite recent efforts, existing Temporal Forgery Localization methodologies exhibit structural vulnerabilities against adaptive adversarial strategies because they fundamentally rely on flawed representational paradigms. Current multimodal approaches typically adopt shallow feature fusion, such as simple feature concatenation or late-stage fusion. Inspired by the foundational Transformer architecture \cite{ref42}, many recent methods attempt to bridge modalities using cross-attention. While these attention-based fusion strategies have achieved remarkable success in general audio-visual tasks, such as weakly-supervised action localization \cite{ref28} and speech recognition \cite{ref29}, their direct application to forensics typically results in merely high-level semantic alignment. This loose coupling mechanism critically ignores the deep intrinsic synchronization inherent in multimodal data, particularly between audio and visual signals. By relying on noise-invariant extractors\cite{ref4, ref5, ref24, ref27, ref43} and late-stage fusion, these methods inadvertently suppress subtle and high-frequency artifacts inherent to neural rendering, such as microscopic audio-visual desynchronization and phase jitter.

Furthermore, directly adapting architectures from temporal action localization \cite{ref19, ref32} introduces a critical bottleneck. Hierarchical downsampling, which is pervasive in temporal action localization models \cite{ref35, ref40}, inherently smooths out the high frequency details that are essential for multimedia forgery forensics \cite{ref36, ref49}. To refine localization boundaries, several approaches employ cascaded inference or multi stage auxiliary supervision \cite{ref5, ref55}. However, without ground truth labels to guide mask generation during the inference phase, these conventional cascaded frameworks are highly vulnerable to initial prediction inaccuracies. This susceptibility inevitably results in substantial distribution shifts and the recursive propagation of errors, which severely degrades the final localization precision.

\begin{figure*}[!t]
  \centering
  \begin{subfigure}{0.5\textwidth}
    \centering
    \includegraphics[width=\linewidth]{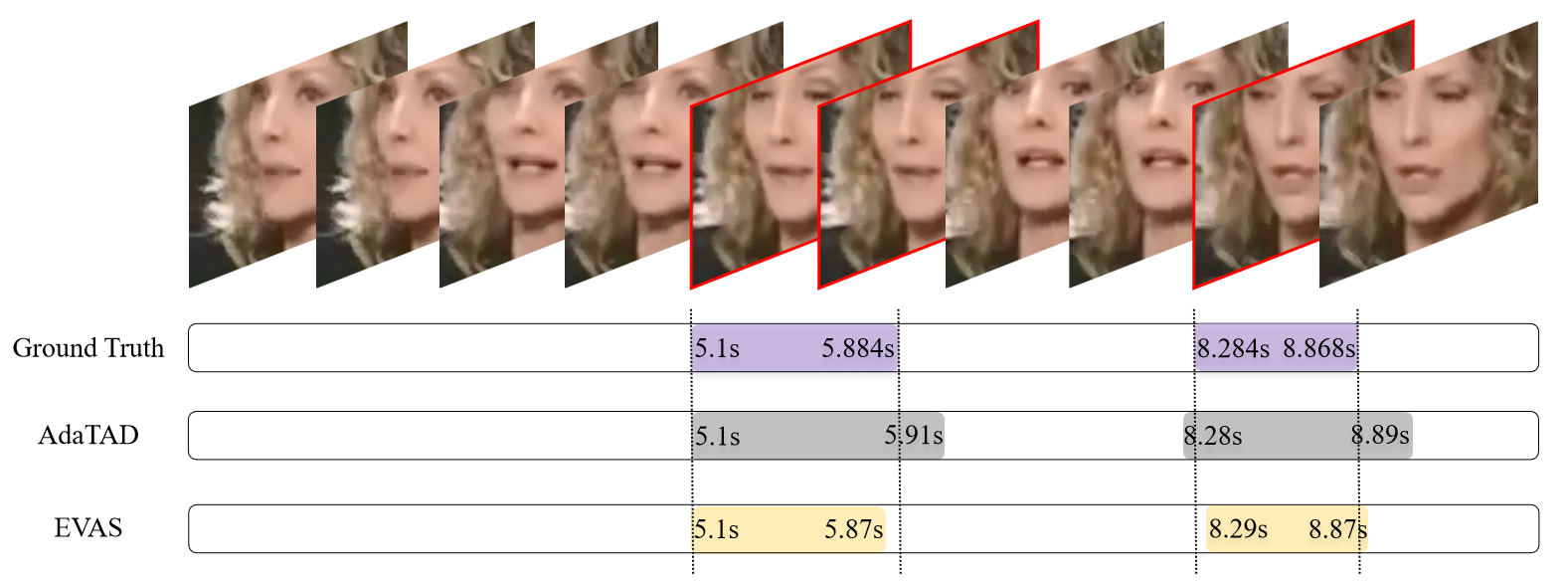}
    \caption{} 
    \label{fig:teaser_a}
  \end{subfigure}
  \hfill
  \begin{subfigure}{0.24\textwidth}
    \centering
    \includegraphics[width=\linewidth]{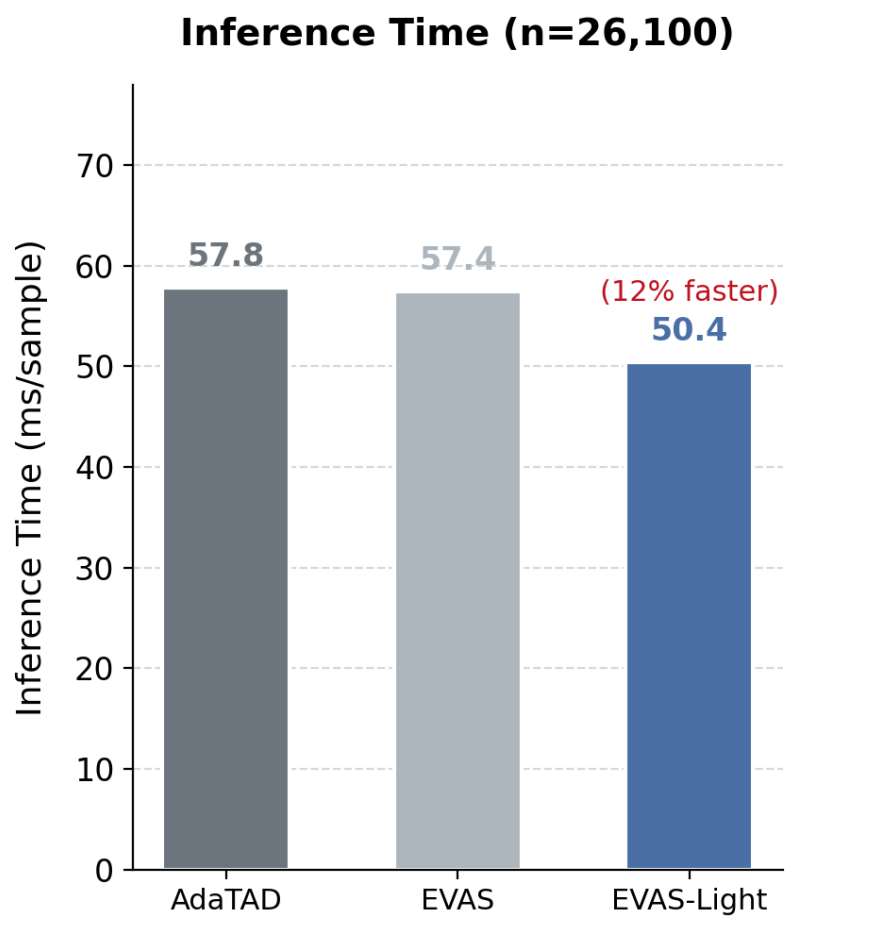}
    \caption{} 
    \label{fig:teaser_b}
  \end{subfigure}
  \hfill
  \begin{subfigure}{0.23\textwidth}
    \centering
    \includegraphics[width=\linewidth]{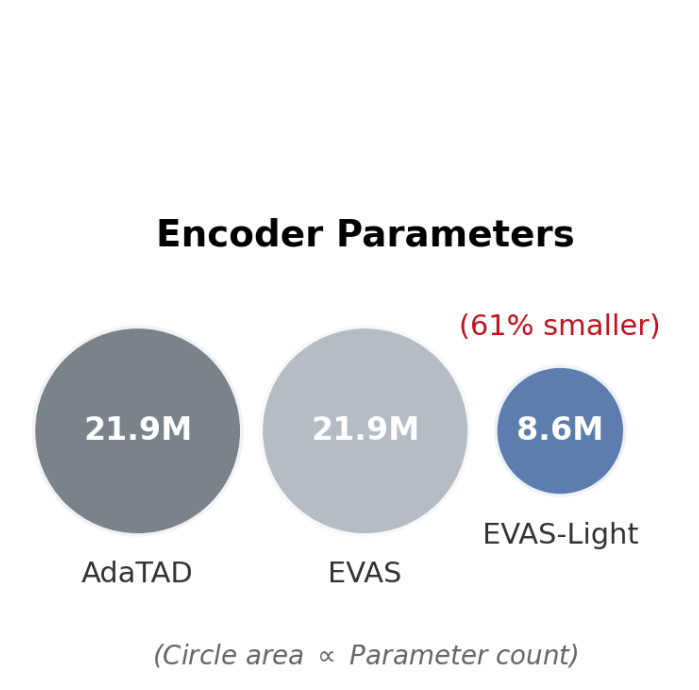}
    \caption{} 
    \label{fig:teaser_c}
  \end{subfigure}

  \caption{(a) The qualitative visualization demonstrates that EVAS aligns more precisely with ground truth boundaries compared to the audio augmented AdaTAD baseline. (b) The inference latency comparison shows that EVAS Light achieves a 12 percent reduction in processing time relative to the standard model. (c) The parameter budget analysis highlights a 61 percent decrease in video encoder parameters for the lightweight variant.}
  \label{fig:teaser}
\end{figure*}

Additionally, the conventional feed-forward networks utilized in transformer-based models impose a massive computational burden \cite{ref34}, severely hindering their deployment in real-time forensics. The prohibitive inference latency makes it impractical to process long-form high-resolution videos efficiently. To overcome this computational bottleneck, a streamlined architectural design is imperative. By replacing standard dense layers with our HourglassFFN, which incorporates dimensionality reduction and second-order nonlinear activation, the computational overhead can be drastically minimized. This strategic modification ensures that the model preserves its robust representation capacity while achieving extreme efficiency for practical applications.

To break these bottlenecks, we propose EVAS, a pioneering end-to-end multimodal framework designed to fundamentally shift the Temporal Forgery Localization paradigm. Unlike disjoint approaches, EVAS inherently binds audio and visual streams from the early stages of feature representation. Specifically, we design the Multi-Stage Audio-Visual Synergy (MAVS) to continuously verify cross-modal consistency and explicitly expose microscopic synchronization artifacts. Furthermore, to overcome cascaded error propagation during boundary localization, we propose Boundary-Aware Refinement (BAR), which utilizes a decoupled training and inference strategy to achieve steered boundary calibration. The main contributions of this work are highlighted by our novel architectural departures from existing methods:

\begin{itemize}[topsep=0pt, partopsep=0pt, itemsep=6pt, parsep=0pt, leftmargin=*]
    \item We propose the EVAS framework featuring the MAVS,  which utilizes dense mutual queries to bind early-stage audio-visual streams, effectively exposing microscopic synchronization artifacts typically missed by shallow fusion.
    \item We introduce the BAR strategy for steered boundary calibration. By employing a decoupled training-inference paradigm, BAR prevents cascaded error propagation and sharpens transition predictions while maintaining real-time efficiency.
    \item We propose a lightweight HourglassFFN architecture integrating dimensionality reduction and second-order nonlinear activation. This design replaces standard dense layers to significantly slash inference latency without compromising the model's robust representation capacity.
    \item Extensive experiments were conducted on three mainstream temporal forgery localization benchmarks (LAV-DF, AV-Deepfake1M, and TVIL). Our method achieved the best average precision and recall across all three datasets, validating its exceptional effectiveness for fine-grained forensics.
\end{itemize}

\section{Related Work}

\subsection{Multimodal Temporal Forgery Localization}
As deepfake fidelity improves, research is rapidly shifting toward Temporal Forgery Localization to provide precise tampering intervals. Early explorations primarily focused on uni-modal analysis \cite{ref53}, rendering them vulnerable to semantic decoupling attacks where only one modality is manipulated. To address this, recent works incorporate multi-modal information to exploit audio-visual inconsistency cues \cite{ref30}. Methods such as Audio-Visual Feature Fusion \cite{ref47}, UMMAFormer \cite{ref36}, and MFMS \cite{ref41} utilize attention mechanisms, while others use high-frequency noise features \cite{ref49}. However, most methods employ shallow fusion strategies like simple feature concatenation. This loose coupling fails to capture the minute audio-visual misalignments inherent in highly realistic neural rendering.

\subsection{Temporal Action Localization in Forensics}
Temporal Action Localization (TAL) constitutes the methodological foundation of this work. Although snippet-level methods \cite{ref7} and graph-based frame-level methods such as G-TAD \cite{ref8} achieved success, recent innovations significantly enhanced the ability to capture temporal dependencies and localize action boundaries \cite{ref50, ref51, ref19, ref20, ref21}. However, these existing methods cannot be directly transferred to forensic tasks. Unlike semantic actions with significant dynamic characteristics such as running, forgery traces typically manifest as subtle artifacts without obvious motion boundaries. Therefore, the feature interaction and localization mechanisms must be specifically reconstructed for digital forensics.

\begin{figure*}[!t]
    \centering
    \begin{minipage}[t]{0.68\textwidth}
        \centering
        \includegraphics[width=\linewidth, height=6.7cm, keepaspectratio]{MyEVAS.pdf}
    \end{minipage} 
    \begin{minipage}[t]{0.31\textwidth}
        \centering
        \raisebox{0.1cm}{\includegraphics[width=\linewidth, height=6.7cm, keepaspectratio]{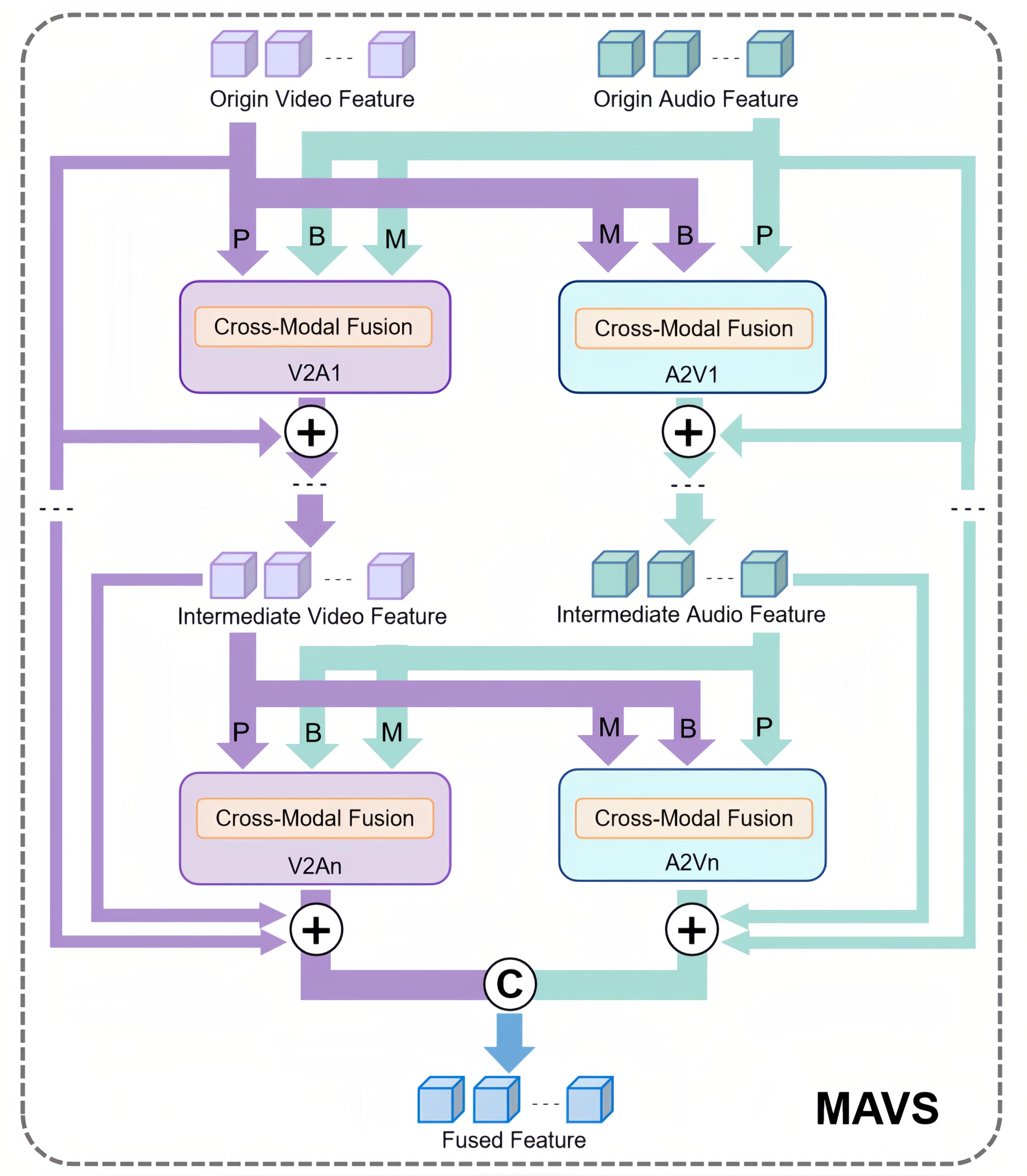}}
    \end{minipage}
    \caption{Overview of the proposed EVAS framework and schematic illustration of the MAVS mechanism. Dashed cubes within the framework represent masked invalid frames, as detailed in Section 3.3. In the MAVS mechanism, arrows labeled $P$, $B$, and $M$ denote the source features for the respective weight matrices in each cross-modal fusion block, while the symbol $C$ indicates the concatenation operation.}
    \label{fig:EVAS_MAVS}
\end{figure*}

\section{Methodology}
\label{sec:methodology}

\subsection{Overview}
Our objective is to robustly detect and localize forgeries in suspect video streams $\mathcal{V}$ under adaptive adversarial attacks. The localized segments can be represented as $\mathcal{G} = \{(s_i, e_i, c_i)\}_{i=1}^{M}$, where $M$ is the number of detected forgery intervals, and $s_i$, $e_i$, and $c_i$ denote the start time, the end time, and the confidence score respectively. To achieve this, $\mathcal{V}$ is formulated as $T$ aligned visual and audio segments $\{v_t\}_{t=1}^{T}$, and a forensic mapping $\mathcal{F}: \mathcal{V} \to \mathcal{G}$ is learned by maximizing the temporal intersection over union with ground truth boundaries to prioritize high-frequency manipulation artifacts over semantic content. Our proposed EVAS framework, illustrated in Figure \ref{fig:EVAS_MAVS}, consists of three core components: a multi-stage audio-visual synergy mechanism utilizing deep mutual queries to expose cross-modal evasion, a boundary-aware refinement module employing progressive auxiliary supervision to counteract boundary confusion, and a lightweight HourglassFFN. By synergizing these specialized modules, EVAS effectively isolates minute synchronization artifacts from pristine background noise. This tailored design ensures high localization accuracy while preserving the computational efficiency requisite for large-scale content moderation.

\subsection{Multi-Stage Audio-Visual Synergy}
\noindent\textbf{Feature Encoding.} \hspace{0.5em}
We preprocess the raw video and decoupled audio streams into standardized tensor formats to establish a foundation for multimodal analysis. Let $\mathcal{X}_{v}$ and $\mathcal{X}_{a}$ denote the raw visual and acoustic inputs. These streams are mapped into continuous latent spaces via specialized feature extraction networks. Specifically, we define the visual feature encoder $\Phi_{v}$ utilizing VideoMAE-S \cite{ref23} equipped with a temporal adapter \cite{ref25}, and the acoustic feature encoder $\Phi_{a}$ employing the BYOL-a network \cite{ref26}. The initial enhanced state representations are formulated as:

\begin{equation}
    \mathbf{H}_{v}^{0} = \Phi_{v}(\mathcal{X}_{v}), \quad \mathbf{H}_{a}^{0} = \Phi_{a}(\mathcal{X}_{a}).
    \label{eq:init_state}
\end{equation}
Here, both the video features $\mathbf{H}_{v}^{0}$ extracted by the video encoder and the audio features $\mathbf{H}_{a}^{0}$ processed by the audio encoder are structured as three-dimensional tensors $\mathbb{R}^{B \times T \times C}$, corresponding to the batch size, temporal dimension, and feature dimension, respectively.

\noindent\textbf{Multi-Stage Audio-Visual Synergy.} \hspace{0.5em}
The representational integrity of feature fusion is a critical determinant of localization accuracy in forensic tasks. Existing methods often rely on rudimentary concatenation or delayed cross-domain mapping, which merely captures macroscopic semantic correlations. Motivated by the necessity for deep integration, we design a specialized mechanism termed Hierarchical State Synchronization. 

After aligning the initial dimensionalities to prevent information loss, the state sequences of these two modalities undergo iterative refinement over $L$ cascaded stages. We articulate the dynamic update rule utilizing the $l$-th layer as a paradigm. For a given target modality, its current state acts as the Probe matrix $\mathbf{P}$, while the complementary modality provides the Basis matrix $\mathbf{B}$ and Message matrix $\mathbf{M}$. Defining the acoustic state update modulated by the visual state, the intermediate variables at layer $l$ are derived via domain-specific linear transformations parametrized by $\mathbf{W}$:

\begin{equation}
    \mathbf{P}_{a}^{l} = \mathbf{H}_{a}^{l}\mathbf{W}_{P}^{l}, \quad \mathbf{B}_{v}^{l} = \mathbf{H}_{v}^{l}\mathbf{W}_{B}^{l}, \quad \mathbf{M}_{v}^{l} = \mathbf{H}_{v}^{l}\mathbf{W}_{M}^{l}.
    \label{eq:linear_proj}
\end{equation}

The cross-modal information transfer function $\Gamma_{a \leftarrow v}^{l}$ is mathematically defined as a non-linear projection over the normalized inner product space:

\begin{equation}
    \Gamma_{a \leftarrow v}^{l} = \text{FFN}\left( \sigma\left( \frac{\mathbf{P}_{a}^{l} (\mathbf{B}_{v}^{l})^\top}{\tau} \right) \mathbf{M}_{v}^{l} \right),
    \label{eq:transfer}
\end{equation}
where $\tau$ is the inherent scaling constant representing the subspace dimensionality and $\sigma$ denotes the softmax normalization operation. The refined state representation $\mathbf{H}_{a}^{l+1}$ is computed through a dense accumulation trajectory:

\begin{equation}
    \mathbf{H}_{a}^{l+1} = \sum_{i=1}^{l} \mathbf{H}_{a}^{i} + \Gamma_{a \leftarrow v}^{l}.
    \label{eq:dense_acc}
\end{equation}

Equation \ref{eq:dense_acc} formalizes the core theoretical postulate of our architecture. By sequentially integrating all preceding historical states, this structural design surpasses the conventional mitigation of signal degradation. Critically, because the Probe vector is strictly synthesized from the target stream, the acoustic modality maintains its inherent topological geometry and semantic dominance. This deep bidirectional alignment strategy guarantees that domain-specific forensic cues remain uncontaminated by heterogeneous visual data, enabling the stable extraction of high-frequency synchronization artifacts. A symmetric formulation is applied to derive the visual state update $\mathbf{H}_{v}^{l+1}$. Following $L$ iterative cycles, where empirical results demonstrate optimal performance at $L=2$, the terminal states are merged via concatenation to formulate the ultimate representation $\mathbf{Z}$ for the detection head:

\begin{equation}
    \mathbf{Z} = \mathbf{H}_{a}^{L} \oplus \mathbf{H}_{v}^{L}.
    \label{eq:final_merge}
\end{equation}

\subsection{Boundary Aware Refinement}
\noindent\textbf{Detector Selection.} \hspace{0.5em}
To effectively localize discontinuous and variable length forged segments, we adopt an anchor free detection architecture \cite{ref35} to detect forged segments. Let $Z \in \mathbb{R}^{T \times C}$ denote the input feature sequence of length $T$ with channel dimension $C$. For a given temporal step $t$, the detector outputs a unified prediction tuple $\hat{y}_t$, which can be formulated as:
\begin{equation}
    \hat{y}_t = (\hat{p}_t, \hat{d}_{s,t}, \hat{d}_{e,t})
\end{equation}
where $\hat{p}_t \in [0, 1]$ represents the classification confidence, and $\hat{d}_{s,t}, \hat{d}_{e,t} \in \mathbb{R}^{+}$ denote the predicted distances to the start and end boundaries respectively. The complete set of predictions for the sequence is denoted as $\hat{Y} = \{\hat{y}_t\}_{t=1}^T = \text{Detector}(Z)$.

\noindent\textbf{Boundary Aware Refinement.} \hspace{0.5em}
In the context of temporal forgery localization, a boundary signifies the critical transition point where the video signal shifts from a pristine state to a manipulated state. To address the subtle high frequency artifacts characterizing forgery boundaries, we propose the Boundary Aware Refinement mechanism. This mechanism implements a progressive auxiliary supervision strategy across $N$ stages, employing a sequence of identical detectors $\text{Detector}_{k}$. The initial input is defined as $F_{m}^{0} = Z$, and subsequent heads serve exclusively to refine the feature space in a cascaded manner.

\noindent\textbf{Iterative Refinement Stages.} \hspace{0.5em}
For the intermediate stage $k$ (where $k \in \{1, 2, \dots, N\}$ and $F_{m}^{0} = Z$), the detection head generates the predicted forgery segment set $\hat{Y}_k = \text{Detector}_k(F_{m}^{k-1})$. Subsequently, to purify the features for the next stage $k+1$, we introduce a strict feature purification mechanism named Mask Invalid Frames. Let $\mathcal{G}$ denote the set of ground truth forgery segments. Based on center sampled points $T_{t} \in \mathcal{G}$, we extract the corresponding predicted temporal bounds $\hat{s}_{k,i}, \hat{e}_{k,i}$. To compensate for prediction jitters, we apply a temporal relaxation strategy. The relaxation factor $\Delta \tau_i$ is defined as a specific percentage $\gamma$ of the predicted temporal segment length $l_{k,i}$. The length is calculated as $l_{k,i} = \hat{e}_{k,i} - \hat{s}_{k,i}$. The relaxation factor is formulated as:
\begin{equation}
    \Delta \tau_i = \gamma \times l_{k,i}
\end{equation}
Empirical hyperparameter analysis demonstrates that the model achieves optimal performance when this specific percentage is set to 0.2. This yields a unified set of refined temporal regions $\mathcal{R}_k$:
\begin{equation}
    \mathcal{R}_k = \bigcup_{i} \left[ \hat{s}_{k,i} - \Delta \tau_i, \hat{e}_{k,i} + \Delta \tau_i \right]
\end{equation}
These regions are then converted into a binary mask vector $M_{k} \in \{0, 1\}^T$ using an indicator function $\mathbb{I}$:
\begin{equation}
    M_{k}[t] = \mathbb{I}(t \in \mathcal{R}_k)
\end{equation}
The mask is applied to the fused features from the previous stage $F_{m}^{k-1}$ via the Hadamard product $\odot$ to generate the input for the next stage:
\begin{equation}
    F_{m}^{k} = F_{m}^{k-1} \odot M_{k}
\end{equation}
By enforcing this operation, the mechanism completely zeros out pristine background information $F_{m}^{k}[t] = \mathbf{0}$ for $t \notin \mathcal{R}_k$. This forces the shared encoder during training to concentrate its gradient updates exclusively on verified forgery contexts. The final stage $N$ generates the ultimate predictions $\hat{Y}_N$ without further masking.

\begin{figure}[!t] 
    \centering
    \includegraphics[width=\linewidth]{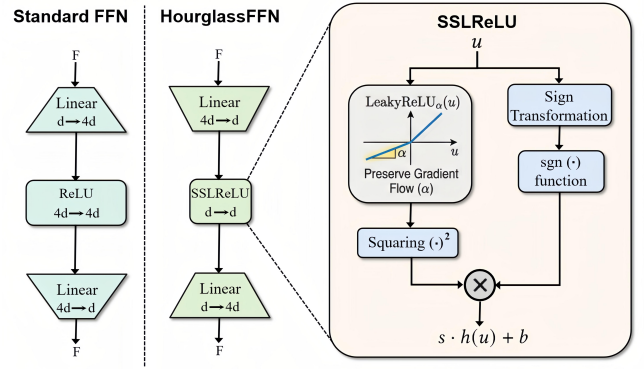} 
    \caption{\textbf{Overview of the HourglassFFN.}} 
    \label{fig:hourglass_ffn} 
\end{figure}

\noindent\textbf{Decoupled Strategy Analysis.} \hspace{0.5em}
The most critical innovation of our refinement mechanism is its explicitly decoupled strategy for training and inference. During training, teacher forcing via ground truth masking guided by $\mathcal{G}$ allows the auxiliary heads to effectively mine hard examples. However, during real world inference, ground truth labels $\mathcal{G}$ are absent, meaning the mask generation must depend recursively on preceding predictions $\tilde{\mathcal{R}}_k \leftarrow f(\hat{Y}_k)$. If initial predictions contain minor localization error $\epsilon$, utilizing them to filter features induces a severe input distribution shift. The recursive masking operation drastically amplifies this error $\epsilon \rightarrow \epsilon'$ triggering catastrophic recursive error propagation and prediction collapse. Therefore, our inference phase discards all auxiliary heads and relies exclusively on the highly robust primary detector:
\begin{equation}
    \hat{Y}_{inference} = \text{Detector}_1(Z)
\end{equation}
This mathematically optimal design choice completely blocks error propagation while ensuring high throughput inference efficiency. The ultimately generated predictions can subsequently be utilized to compute the regression loss.

\subsection{HourglassFFN}
\label{sec:hourglass_ffn}
To enable lightweight deployment, we propose HourglassFFN which aggressively compresses intermediate features from dimension $d$ to $h \ll d$. Standard activations discard crucial information within this narrow bottleneck.

\begin{table*}[!t]
\centering
\caption{Comparison of the performance of various methods on the Lav-DF dataset. Bold values denote the best performance, underlined values denote the second-best performance, and the E2E column indicates whether the model is trained in an end-to-end manner.}
\begingroup
\footnotesize
\setlength{\tabcolsep}{3pt}
\renewcommand{\arraystretch}{1.06}
\begin{tabularx}{\textwidth}{>{\centering}p{2.5cm} >{\centering}p{2.6cm} | >{\centering}p{0.7cm} | c c c c c c c c}
\hline
Methods & Features & E2E & AP$@$0.5 & AP$@$0.75 & AP$@$0.95 & mAP & AR$@$10 & AR$@$20 & AR$@$50 & AR$@$100 \\
\hline
MDS\cite{ref30} & Visual & $\times$ & 12.78 & 1.62 & 0.00 & 4.80 & 37.88 & 36.71 & 34.39 & 32.15 \\
AGT\cite{ref31} & Visual & $\times$ & 17.85 & 9.42 & 0.11 & 9.13 & 43.15 & 34.23 & 24.59 & 16.71 \\
BMN\cite{ref32} & Visual & $\times$ & 24.01 & 7.61 & 0.07 & 10.56 & 53.26 & 41.24 & 31.60 & 26.93 \\
BMN (I3D)\cite{ref32} & Visual & $\times$ & 10.56 & 1.66 & 0.00 & 4.07 & 48.49 & 44.39 & 37.13 & 31.55 \\
AVFusion\cite{ref33} & Visual+Audio & $\times$ & 65.38 & 23.89 & 0.11 & 29.79 & 62.98 & 59.26 & 54.80 & 52.11 \\
\multirow{2}{*}{BA-TFD~\cite{ref6}}
 & Visual & $\times$ & 58.55 & 28.60 & 0.16 & 29.10 & 62.49 & 58.77 & 53.86 & 50.29 \\
 & Visual+Audio & $\times$ & 76.90 & 38.50 & 0.25 & 38.55 & 66.90 & 64.08 & 60.77 & 58.42 \\
ActionFormer\cite{ref35} & Visual & $\times$ & 95.34 & 90.20 & 23.73 & 69.76 & 88.41 & 89.63 & 90.33 & 90.41 \\
\multirow{2}{*}{UMMAFormer~\cite{ref36}}
 & Visual & $\times$ & 97.30 & 92.96 & 25.68 & 71.98 & 90.19 & 90.85 & 91.14 & 91.18 \\
 & Visual+Audio & $\times$ & \textbf{98.83} & \underline{95.54} & 37.61 & 77.33 & 92.10 & 92.42 & 92.47 & 92.48 \\
DiMoDif\cite{ref37} & Visual+Audio & $\times$ & 95.50 & 87.90 & 20.60 & 68.00 & 91.40 & 92.70 & 93.70 & 94.20 \\
TriDet\cite{ref40} & Visual+Audio & $\times$ & 96.29 & 86.84 & 23.64 & 68.92 & 88.69 & 89.71 & 90.39 & 91.00 \\
MFMS\cite{ref41} & Visual+Audio & $\times$ & \underline{98.47} & 94.15 & 27.80 & 73.47 & 90.02 & 90.46 & 90.65 & 90.69 \\
\hline
EVAS & Visual+Audio & $\checkmark$ & 98.41 & \textbf{96.61} & \textbf{88.63} & \textbf{94.55} & \textbf{97.59} & \textbf{97.67} & \textbf{97.73} & \textbf{97.81} \\
EVAS (Lightweight) & Visual+Audio & $\checkmark$ & 96.25 & 92.40 & \underline{81.13} & \underline{89.93} & \underline{95.71} & \underline{95.92} & \underline{96.12} & \underline{96.28} \\
\hline
\end{tabularx}
\endgroup
\label{tab:comparison_Lavdf}
\end{table*}

\begin{table*}[!t]
\centering
\caption{Comparison of the performance of various methods on the AV-Deepfake1M dataset. Bold values denote the best performance, underlined values denote the second-best performance, and the E2E column indicates whether the model is trained in an end-to-end manner.}
\begingroup
\footnotesize
\setlength{\tabcolsep}{4pt}
\renewcommand{\arraystretch}{1.06}
\begin{tabularx}{\textwidth}{>{\centering\arraybackslash}p{3.5cm} >{\centering\arraybackslash}p{2.6cm} | >{\centering\arraybackslash}p{0.7cm} | c c c c c c c}
\hline
Methods & Features & E2E & AP$@$0.5 & AP$@$0.75 & AP$@$0.95 & mAP & AR$@$10 & AR$@$20 & AR$@$50 \\
\hline
MesoInception4\cite{ref38} & Visual & $\times$ & 8.50 & 5.16 & 0.50 & 4.72 & 35.78 & 39.00 & 39.27 \\
ActionFormer\cite{ref35} & Visual & $\times$ & 20.24 & 5.73 & 0.07 & 8.68 & 19.11 & 19.81 & 19.97 \\
BA-TFD\cite{ref6} & Visual+Audio & $\times$ & 37.37 & 6.34 & 0.02 & 14.58 & 30.66 & 35.95 & 45.55 \\
BA-TFD+\cite{ref39} & Visual+Audio & $\times$ & 44.42 & 13.64 & 0.03 & 19.36 & 34.67 & 40.37 & 48.86 \\
UMMAFormer\cite{ref36} & Visual+Audio & $\times$ & 51.64 & 28.07 & 1.58 & 27.10 & 42.09 & 43.45 & 44.07 \\
DiMoDif\cite{ref37} & Visual+Audio & $\times$ & \underline{86.93} & \underline{75.95} & \underline{5.43} & \underline{56.10} & \underline{78.84} & \underline{80.25} & \underline{81.57} \\
\hline
EVAS & Visual+Audio & $\checkmark$ & \textbf{90.53} & \textbf{85.10} & \textbf{34.96} & \textbf{70.20} & \textbf{82.88} & \textbf{83.44} & \textbf{83.89} \\
\hline
\end{tabularx}
\endgroup
\label{tab:comparison_AVDeepfake1M}
\end{table*}

\noindent\textbf{SSLReLU.  } \hspace{0.5em}
We introduce Self Stabilizing Leaky ReLU or SSLReLU to maximize information flow. It preserves gradient flow via a negative slope $\alpha$ and locally enforces zero mean and unit variance constraints to mitigate gradient vanishing. The activation is defined as:
\begin{equation}
    \Phi(u) = \text{SSLReLU}(u) = s \cdot h(u) + b,
\end{equation}
where $h(u) = \text{sgn}(u) \cdot (\text{LeakyReLU}_{\alpha}(u))^2$ is the signed squared transformation. Parameters $s$ and $b$ stabilize signal propagation.

\noindent\textbf{Normalization Constants. } \hspace{0.5em}
Assuming the normalized input follows a standard normal distribution $u \sim \mathcal{N}(0, 1)$, we derive closed form solutions for $s$ and $b$. Using Gaussian symmetry where the contribution of each half space to the $k$-th moment is $\frac{1}{2}\mu_k$ with $\mu_2=1$ and $\mu_4=3$, the moments of $h(u)$ are calculated as:

\begin{align}
    \mathbb{E}[h(u)] &= \underbrace{\frac{1}{2}\mu_2}_{\text{positive}} + \underbrace{\frac{1}{2}(-\alpha^2)\mu_2}_{\text{negative}} = \frac{1 - \alpha^2}{2}, \\
    \mathbb{E}[h(u)^2] &= \underbrace{\frac{1}{2}\mu_4}_{\text{positive}} + \underbrace{\frac{1}{2}(-\alpha^2)^2\mu_4}_{\text{negative}} = \frac{3(1 + \alpha^4)}{2}.
\end{align}

To strictly enforce the unit variance property, the normalization constants are determined by:

\begin{equation}
    s = \frac{1}{\sqrt{\mathbb{E}[h(u)^2] - (\mathbb{E}[h(u)])^2}}, \quad b = -s \cdot \mathbb{E}[h(u)].
\end{equation}

These constants counteract the distribution shift.

\begin{table}[!t]
\centering
\caption{Efficiency Comparison}
\label{tab:efficiency_comparison}
\small
\setlength{\tabcolsep}{9pt}
\renewcommand{\arraystretch}{1.08}
\begin{tabular}{cc}
\hline
Methods & Inference Time (ms) \\
\hline
UMMAFormer~\cite{ref36} & 29850 \\
EVAS & 57 \\
EVAS (Lightweight) & \textbf{50} \\
\hline
\end{tabular}
\end{table}

\noindent\textbf{Efficiency Analysis.} \hspace{0.5em}
We evaluate the efficiency of our approach under strictly identical experimental settings. Experimental results demonstrate that our lightweight EVAS method requires 
only 50ms for the inference of a single video, whereas the UMMAFormer baseline (including the feature extraction component) takes 29850ms. Furthermore, at the module level, by substituting the wide-expansion FFN with our proposed lightweight HourglassFFN architecture, the total inference time on the validation set is reduced from 1499s to 1316s. This corresponds to a latency reduction of approximately 12.2\%. These results demonstrate that the efficiency gains derived from reduced memory access and parameter count effectively outweigh the computational overhead introduced by the activation function.

\begin{table*}[!t]
\centering
\caption{Comparison of the performance of various methods on the TVIL dataset. Bold values denote the best performance, underlined values denote the second-best performance, and the E2E column indicates whether the model is trained in an end-to-end manner.}
\begingroup
\footnotesize
\setlength{\tabcolsep}{4pt}
\renewcommand{\arraystretch}{1.06}
\begin{tabularx}{\textwidth}{>{\centering\arraybackslash}p{3.0cm} >{\centering\arraybackslash}p{1.6cm} | >{\centering\arraybackslash}p{0.7cm} | c c c c c c c c}
\hline
Methods & Features & E2E & AP$@$0.5 & AP$@$0.75 & AP$@$0.95 & mAP & AR$@$10 & AR$@$20 & AR$@$50 & AR$@$100 \\
\hline
TAGS\cite{ref54} & Visual & $\times$ & 18.40 & 12.68 & 0.09 & 10.39 & 24.41 & 25.05 & 25.56 & 25.56 \\
DCAN\cite{ref55} & Visual & $\times$ & 82.75 & 75.00 & 3.22 & 53.66 & 64.73 & 66.02 & 68.82 & 69.97 \\
ActionFormer\cite{ref35} & Visual & $\times$ & 86.27 & 83.03 & 28.17 & 65.82 & 84.82 & 85.77 & 88.10 & 88.49 \\
UMMAFormer\cite{ref36} & Visual & $\times$ & \underline{88.68} & \underline{84.70} & \underline{62.43} & \underline{78.60} & \underline{87.09} & \underline{88.21} & \underline{90.43} & \underline{91.16} \\
\hline
EVAS & Visual & $\checkmark$ & \textbf{94.91} & \textbf{93.16} & \textbf{62.56} & \textbf{83.54} & \textbf{91.96} & \textbf{92.39} & \textbf{93.25} & \textbf{94.03} \\
\hline
\end{tabularx}
\endgroup
\label{tab:comparison_TVIL}
\end{table*}

\begin{table*}[!t]
    \centering
    \caption{Ablation study on the effectiveness of key components. A+V: Audio-Visual Baseline; BAR: Boundary-Aware Refinement; MAVS: Multi-Stage Audio-Visual Synergy.}
    \resizebox{0.85\linewidth}{!}{%
    \renewcommand{\arraystretch}{1.12}
    \begin{tabular}{ccc | cccc | cccc}
        \hline
        A+V & BAR & MAVS & AP$@$0.5 & AP$@$0.75 & AP$@$0.95 & mAP & AR$@$10 & AR$@$20 & AR$@$50 & AR$@$100 \\
        \hline
        \checkmark & \xmark     & \xmark     & 97.78 & 95.83 & 83.96 & 92.52 & 96.40 & 96.54 & 96.63 & 96.73 \\
        \checkmark & \checkmark & \xmark     & \textbf{98.68} & \textbf{96.95} & 88.25 & \textbf{94.63} & 97.56 & 97.63 & 97.71 & 97.78 \\
        \checkmark & \checkmark & \checkmark & 98.41 & 96.61 & \textbf{88.63} & 94.55 & \textbf{97.59} & \textbf{97.67} & \textbf{97.73} & \textbf{97.81} \\
        \hline
    \end{tabular}%
    }
    \label{tab:ablation_components}
\end{table*}

\begin{table*}[!t]
    \centering
    \caption{Analysis of the number of attention layers in the Multi-Stage Audio-Visual Synergy module.}
    \resizebox{0.8\linewidth}{!}{%
    \renewcommand{\arraystretch}{1.12}
    \begin{tabular}{c| cccc | cccc}
        \hline
        Layers ($L$) & AP$@$0.5 & AP$@$0.75 & AP$@$0.95 & mAP & AR$@$10 & AR$@$20 & AR$@$50 & AR$@$100 \\
        \hline
        $0$ & \textbf{98.68} & \textbf{96.95} & 88.25 & \textbf{94.63} & 97.56 & 97.63 & 97.71 & 97.78 \\
        $1$ & 97.53 & 95.27 & 77.43 & 90.08 & 95.15 & 95.32 & 95.48 & 95.62 \\
        $2$ & 98.41 & 96.61 & \textbf{88.63} & 94.55 & \textbf{97.59} & \textbf{97.67} & \textbf{97.73} & \textbf{97.81} \\
        $3$ & 97.34 & 95.26 & 76.66 & 89.75 & 95.02 & 95.16 & 95.31 & 95.44 \\
        \hline
    \end{tabular}%
    }
    \label{tab:ablation_MAVSlayers}
\end{table*}

\subsection{Inference and Loss}
During the inference phase, we strictly adhere to the decoupled training inference strategy detailed in Section 4.3. The training objective incorporates both classification and boundary regression tasks, utilizing Focal Loss \cite{ref45} to address class imbalance and Distance IoU Loss \cite{ref44} for precise boundary localization. We compute the aggregate loss by averaging the classification and regression losses across the three detection heads. The total objective function is defined as:
\begin{align}
\mathcal{L}_{total} = \frac{1}{T} \sum_{t=1}^{T} \overline{\mathcal{L}}_{cls}(t) + \frac{\lambda}{N_{pos}} \sum_{t=1}^{T} \mathbb{I}_{t} \cdot \overline{\mathcal{L}}_{reg}(t),
\end{align}

where $T$ represents the temporal sequence length, and $N_{pos}$ denotes the total count of positive or forged samples within the batch. The indicator function equals 1 if the time step falls within a forged segment, and 0 otherwise. The hyperparameter $\lambda$ balances the contribution of the regression term. This loss is computed across all feature pyramid levels and averaged over the training batch.

\section{Experiments}

This section introduces the datasets and experimental details followed by a comparison of EVAS with state of the art methods. Subsequently, we conduct a series of ablation studies to verify the effectiveness of each component of the model.
\vspace{0.3em}

\noindent\textbf{Datasets.} \hspace{0.2em} 
To comprehensively evaluate the effectiveness and robustness of our proposed framework, we conducted experiments on three challenging benchmarks namely Lav-DF \cite{ref6}, AV-deepfake1M \cite{ref56}, and TVIL \cite{ref36}.

The Localized Audio Visual DeepFake benchmark Lav-DF along with AV-deepfake1M and TVIL are large scale datasets specifically designed for the task of multimodal temporal forgery localization. These datasets are characterized by variable length manipulated segments that are typically short and tightly interwoven with pristine content. Specifically, the duration of forged segments accounts for only a small proportion of the total video length making the forged content difficult to detect. Furthermore, the datasets cover diverse forgery types including face swapping and voice conversion which pose significant challenges for detecting high frequency artifacts and audio visual inconsistencies. To ensure a rigorous and standardized assessment, all subsequent experiments in this work are evaluated using the official full sets of these benchmarks. Adhering to this standard protocol guarantees that our reported metrics are directly comparable with existing state of the art baselines thereby validating the robustness and generalization capability of our proposed method.
\vspace{0.3em}

\noindent\textbf{Evaluation Metrics.} \hspace{0.2em} This study follows the standard evaluation protocol for temporal forgery localization~\cite{ref36} using Intersection over Union thresholds 0.5, 0.75, and 0.95 to calculate Average Precision for a comprehensive assessment of model performance. For recall calculation, temporal Intersection over Union thresholds are set to 0.5, 0.75, and 0.95, and the Average Recall is computed under the condition of candidate segment numbers selected from 10, 20, 50, and 100. Finally, the average recall across the three temporal Intersection over Union thresholds is taken as the comprehensive evaluation metric.
\vspace{0.3em}

\begin{table*}[!t]
    \centering
    \caption{Analysis of the number of detection heads during training.}
    \resizebox{0.8\linewidth}{!}{%
    \renewcommand{\arraystretch}{1.12}
    \begin{tabular}{c| cccc | cccc}
        \hline
        Stages & AP$@$0.5 & AP$@$0.75 & AP$@$0.95 & mAP & AR$@$10 & AR$@$20 & AR$@$50 & AR$@$100 \\
        \hline
        $1$ & 97.66 & 95.78 & 84.10 & 92.51 & 96.36 & 96.49 & 96.59 & 96.71 \\
        $2$ & 97.24 & 95.01 & 74.72 & 88.99 & 94.61 & 94.87 & 95.03 & 95.19 \\
        $3$ & \textbf{98.41} & \textbf{96.61} & \textbf{88.63} & \textbf{94.55} & \textbf{97.59} & \textbf{97.67} & \textbf{97.73} & \textbf{97.81} \\
        $4$ & 97.17 & 94.88 & 75.29 & 89.11 & 94.67 & 94.93 & 95.11 & 95.24 \\
        \hline
    \end{tabular}%
    }
    \label{tab:ablation_aux_heads}
\end{table*}

\begin{figure*}[!t]
    \centering
    
    \begin{minipage}[b]{0.48\linewidth}
        \centering
        \includegraphics[width=0.71\linewidth]{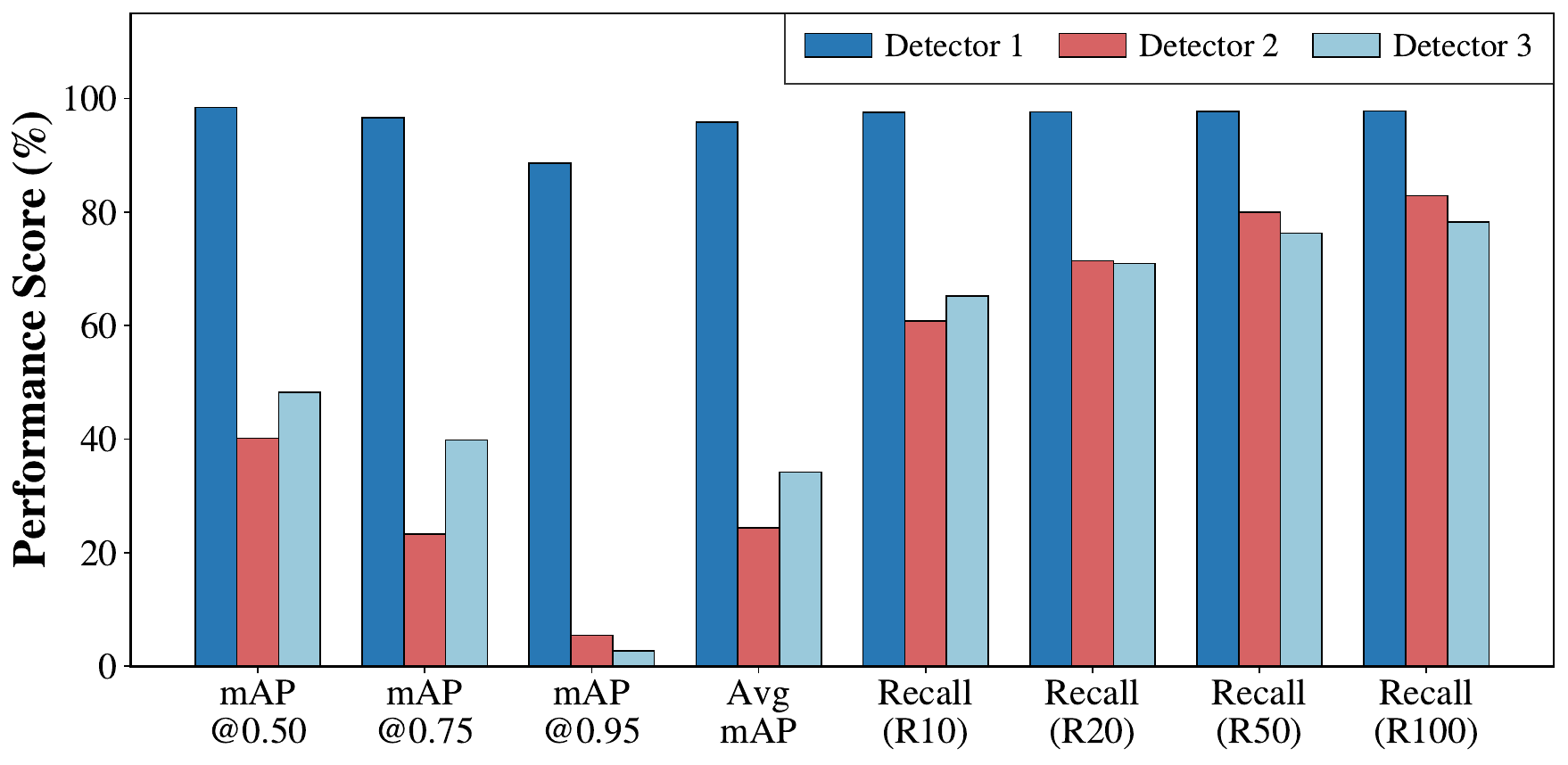} 
        \caption{Analysis of model inference performance using detection heads from different stages.}
        \label{fig:ablation_stage}
    \end{minipage}
    \hfill 
    \begin{minipage}[b]{0.48\linewidth}
        \centering
        \includegraphics[width=0.95\linewidth]{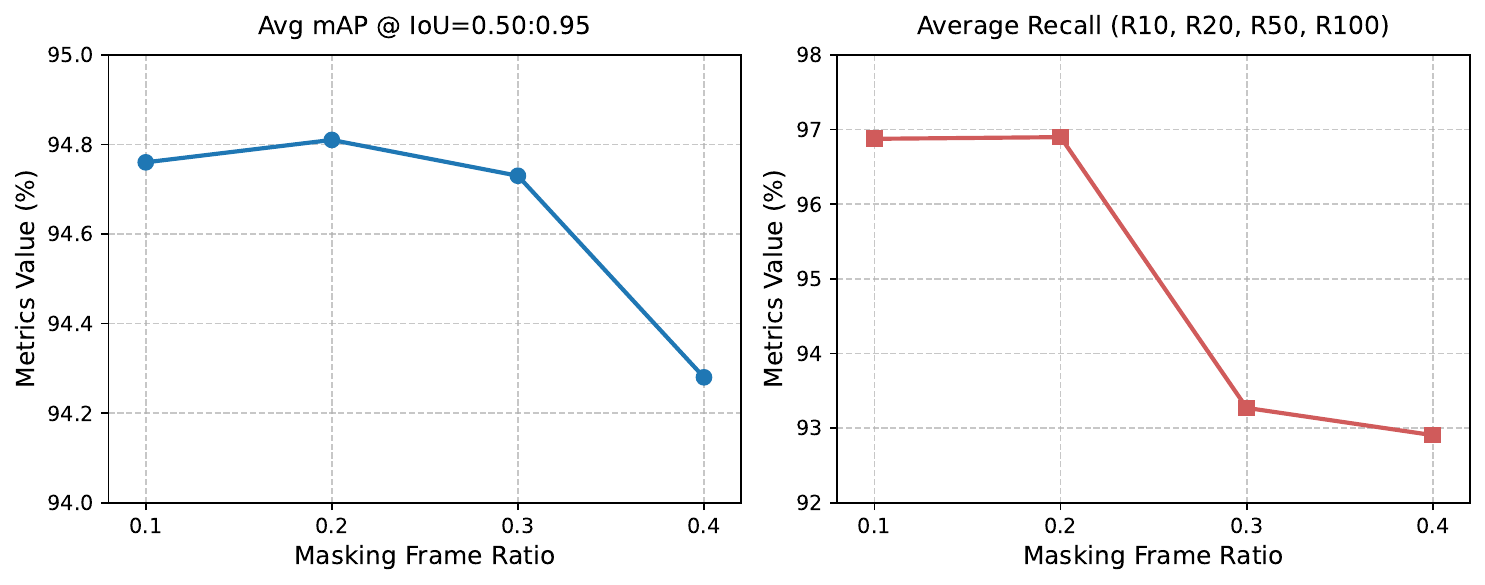} 
        \caption{Analysis of model performance under different masking frame ratios on the Lav-DF dataset.}
        \label{fig:masking_ratio}
    \end{minipage}
\end{figure*}

\begin{figure*}[!t]
    \centering
    
    \includegraphics[width=\textwidth]{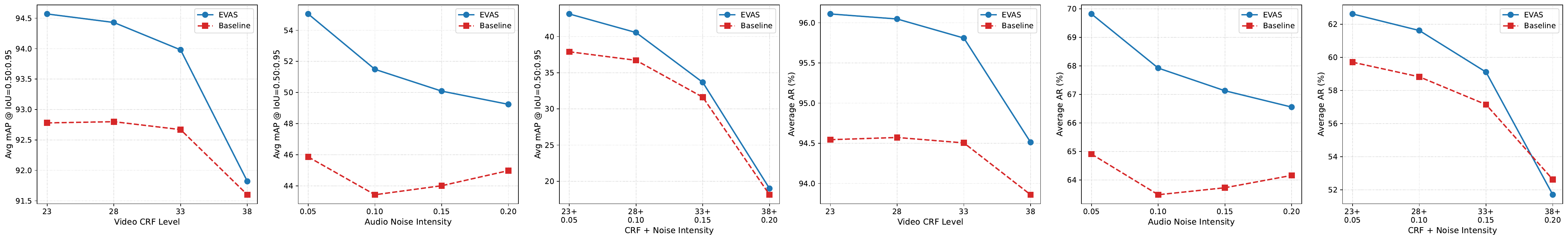}
    
    \caption{Analysis of model robustness under varying video compression and audio noise intensities on the LAVDF dataset.}
    \label{fig:robustness_experiment}
\end{figure*}

\begin{figure*}[!t]
    \centering
    \includegraphics[width=0.85\linewidth]{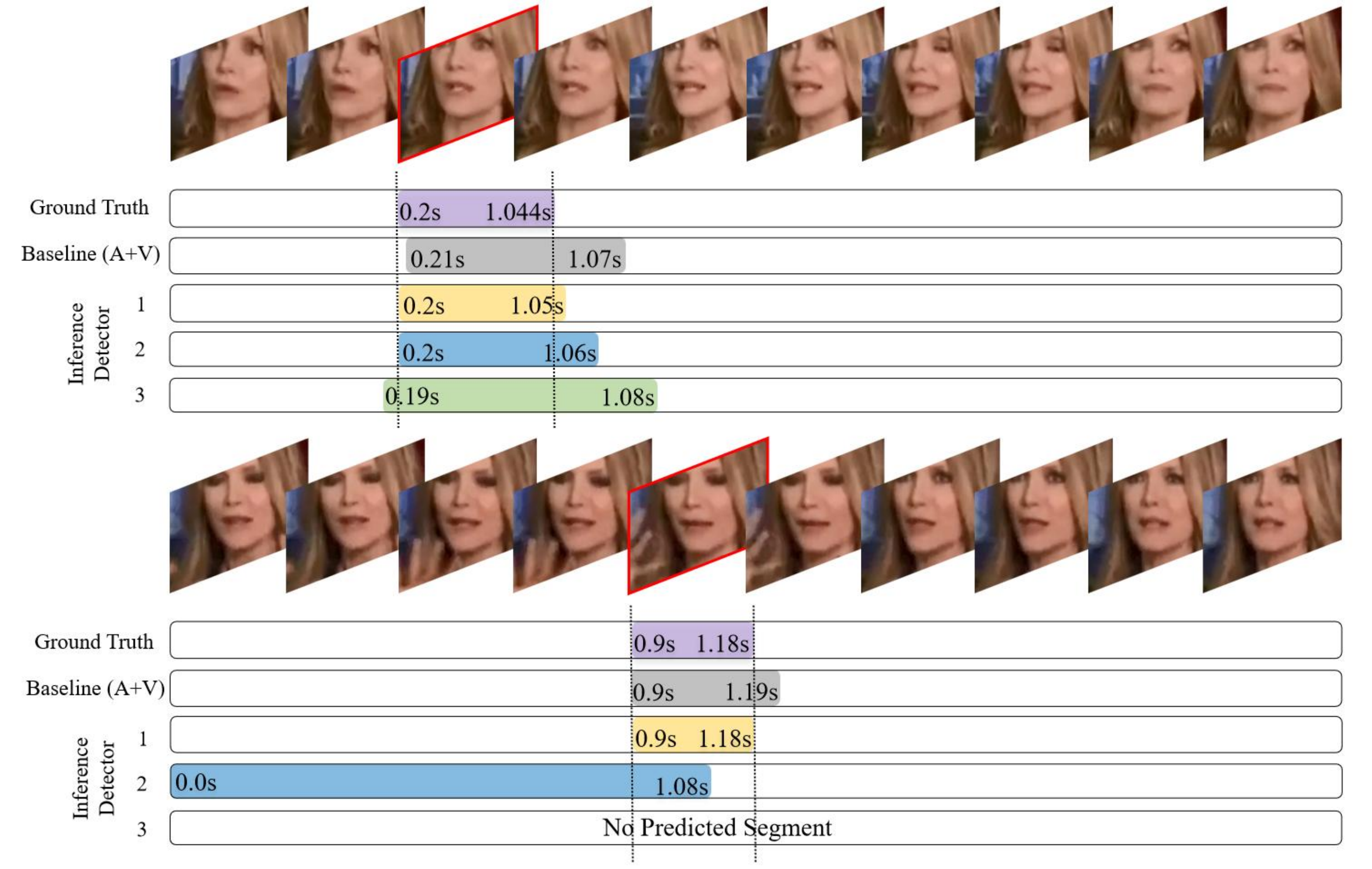}
    \caption{\textbf{Visualization of localization results on the Lav-DF benchmark. We compare the ground truth shown in purple with predictions from the baseline shown in gray and the three sequential detection heads of EVAS. Detector1, Detector2, and Detector3 correspond to the inference outputs utilizing the first, second, and third detection heads respectively. The red borders in the frame sequence denote the manipulated regions.}}
    \label{fig:stacked_vis}
\end{figure*}

\noindent\textbf{Training Settings.} \hspace{0.2em} We optimize the network using the AdamW optimizer with a weight decay of 0.05. The initial learning rate is set to 1e-3 for the detection head and 1e-4 for the tunable modules within the frozen backbone. The learning rate is modulated by a cosine annealing scheduler with a linear warmup. To mitigate overfitting, an early stopping mechanism is applied, terminating training at a maximum of 12 epochs. The batch size is set to 16 per GPU, and the sampling window for each video is 768. Gradient clipping with a maximum norm of 1.0 is utilized to stabilize the training process. For feature extraction, the audio sampling rate is 16 kHz, yielding output feature dimensions of 3072 and 384 for the audio and video encoders, respectively. The proposed method is implemented using PyTorch and the MMAction2 framework, and all experiments are conducted on four NVIDIA V100 GPUs.

\subsection{Comparison Experiments}
In this subsection, we systematically compare our proposed framework against state-of-the-art temporal forgery localization methods. The quantitative outcomes for the Lav-DF dataset are detailed in Table~\ref{tab:comparison_Lavdf}. Our primary model, EVAS, establishes a new performance standard and dominates the leaderboard across all critical metrics. Under the most stringent localization threshold of average precision at 0.95 ($AP@0.95$), the full EVAS model achieves a remarkable score of 88.63\%. This represents a massive improvement over the best existing competitor, UMMAFormer, which attains 37.61\%. This decisive advantage empirically validates that while conventional methods struggle to delineate precise forgery boundaries, our boundary-aware refinement mechanism effectively isolates fine-grained high-frequency artifacts at manipulation transition points. Furthermore, EVAS demonstrates unparalleled recall capabilities by reaching an average recall at 100 ($AR@100$) of 97.81\%, which guarantees minimal missed detections in security-critical scenarios.

Crucially, this superior localization accuracy does not strictly require extensive computational overhead. Our parameter-efficient variant, EVAS Lightweight, secures the second-best overall performance by yielding an $AP@0.95$ of 81.13\%. This configuration still eclipses all prior baseline architectures, proving that our multi-stage audio-visual synergy framework maintains exceptional cross-modal representational integrity under constrained parameter budgets.

We also extend our comparative analysis to the AV-Deepfake1M and TVIL datasets, with results presented in Table~\ref{tab:comparison_AVDeepfake1M} and Table~\ref{tab:comparison_TVIL} respectively. On the challenging AV-Deepfake1M benchmark, EVAS achieves a mean average precision (mAP) of 70.20\%, significantly outperforming alternative methods and demonstrating the robustness of our hierarchical state synchronization mechanism. The TVIL evaluation additionally confirms that our architecture remains highly effective even for unimodal visual inputs. By thoroughly blocking error propagation through our decoupled inference strategy, EVAS reaches a top mAP of 83.54\%, reinforcing the universal applicability of our cascaded refinement paradigm.

\subsection{Ablation Experiments}
Table~\ref{tab:ablation_components} summarizes the ablation studies validating our core components. Integrating the boundary aware refinement module increases the baseline mean average precision from 92.52\% to 94.63\% by concentrating training on verified forgery contexts. The subsequent addition of the multi stage audio visual synergy mechanism further elevates the average precision at 0.95 to 88.63\% and average recall at 100 to 97.81\%. Furthermore, Figure~\ref{fig:ablation_stage} demonstrates the critical necessity of our decoupled inference strategy. Relying exclusively on the primary detector prevents catastrophic performance degradation and error cascades caused by recursive masking operations in auxiliary heads.

\subsection{Parameter Sensitivity Analysis}
In this subsection, we systematically investigate the influence of critical hyperparameters on our proposed framework. Table~\ref{tab:ablation_MAVSlayers} details the effect of varying the number of hierarchical attention layers within the multi stage audio visual synergy module. Configuring this parameter to two layers yields the highest strict localization accuracy, achieving an average precision at 0.95 of 88.63\%. Additionally, Table~\ref{tab:ablation_aux_heads} presents an evaluation of different cascaded refinement stages during the training phase. The three stage architecture delivers the optimal balance, maximizing the mean average precision at 94.55\%. Furthermore, Figure~\ref{fig:masking_ratio} illustrates the performance variations caused by the temporal relaxation factor used for masking invalid frames. A masking ratio of 0.2 represents the optimal threshold, generating a peak mean average precision of 94.81\% and an average recall of 96.90\%. Ratios exceeding this value inadvertently incorporate excessive background noise, which steadily degrades the boundary detection capability of the network.

\subsection{Robustness Analysis}
We evaluate the robustness of our EVAS framework, optimized on the clean LAVDF dataset, against environmental interferences including Constant Rate Factor (CRF) video compression and audio white noise. In this evaluation, the baseline refers to the Audio-Visual (A+V) model detailed in Table \ref{tab:ablation_components}. Figure \ref{fig:robustness_experiment} demonstrates that while performance naturally degrades as the video CRF increases from 23 to 38 or the audio noise scales from 0.05 to 0.20, EVAS consistently outperforms the baseline in both mean Average Precision (mAP) and Average Recall (AR) across most noise settings. For example, at the maximum 0.20 audio noise intensity, EVAS maintains an average recall exceeding 66.5\%. This superior robustness is attributed to the Multi-Stage Audio-Visual Synergy (MAVS) module isolating minute synchronization artifacts, alongside the Boundary-Aware Refinement (BAR) module which mitigates boundary confusion under severe signal corruption.

\subsection{Localization Visualization Analysis}
In this subsection, we qualitatively evaluate localization performance using Figure~\ref{fig:stacked_vis}. The visual evidence illustrates that the primary detector achieves highly precise boundary alignment. Furthermore, it demonstrates that employing subsequent cascaded heads during inference triggers severe error propagation and prediction collapse, completely validating our decoupled inference strategy.

\section{Conclusion}
In this paper, we propose EVAS, an end-to-end temporal forgery localization framework. Our multi-stage audio-visual synergy mechanism ensures deep cross-modal verification. Furthermore, our boundary-aware refinement module employs a decoupled strategy to capture high-frequency artifacts during training while preventing inference error propagation. Supported by a lightweight architecture, extensive evaluations demonstrate that EVAS achieves state-of-the-art localization performance.

\FloatBarrier


\begin{thebibliography}{56}

\bibitem{ref1}
Fanxiao Li, Jiaying Wu, Tingchao Fu, Yunyun Dong, Bingbing Song, Wei Zhou. 2025.
\textit{Drifting Away from Truth: GenAI-Driven News Diversity Challenges LVLM-Based Misinformation Detection}.
\url{https://doi.org/10.48550/arXiv.2508.12711}

\bibitem{ref2}
Lvpan Cai, Haowei Wang, Jiayi Ji, Yanshu Zhoumen, Shen Chen, Taiping Yao, Xiaoshuai Sun. 2025.
\textit{Zooming In on Fakes: A Novel Dataset for Localized AI-Generated Image Detection with Forgery Amplification Approach}.
\url{https://doi.org/10.48550/arXiv.2504.11922}

\bibitem{ref3}
Tong Qiao, Shichuang Xie, Yanli Chen, Florent Retraint, Xiangyang Luo. 2024.
\textit{Fully Unsupervised Deepfake Video Detection Via Enhanced Contrastive Learning}.
\url{https://doi.org/10.1109/TPAMI.2024.3356814}

\bibitem{ref4}
Christian Szegedy, Wei Liu, Yangqing Jia, Pierre Sermanet, Scott Reed, Dragomir Anguelov, Dumitru Erhan, Vincent Vanhoucke, Andrew Rabinovich. 2015.
\textit{Going deeper with convolutions}.
\url{https://doi.org/10.1109/CVPR.2015.7298594}

\bibitem{ref5}
Zhaowei Cai, Nuno Vasconcelos. 2018.
\textit{Cascade R-CNN: Delving into High Quality Object Detection}.
\url{https://doi.org/10.1109/CVPR.2018.00644}

\bibitem{ref6}
Zhixi Cai, Kalin Stefanov, Abhinav Dhall, Munawar Hayat. 2022.
\textit{Do You Really Mean That? Content Driven Audio-Visual Deepfake Dataset and Multimodal Method for Temporal Forgery Localization}.
In 2022 International Conference on Digital Image Computing: Techniques and Applications (DICTA), 1–10.
\url{https://doi.org/10.1109/DICTA56598.2022.10034605}

\bibitem{ref7}
Shaoqing Ren, Kaiming He, Ross Girshick, Jian Sun. 2016.
\textit{Faster R-CNN: Towards Real-Time Object Detection with Region Proposal Networks}.
\url{https://doi.org/10.1109/TPAMI.2016.2577031}

\bibitem{ref8}
Mengmeng Xu, Chen Zhao, David S. Rojas, Ali Thabet, Bernard Ghanem. 2020.
\textit{G-TAD: Sub-Graph Localization for Temporal Action Detection}.
\url{https://doi.org/10.1109/CVPR42600.2020.01017}

\bibitem{ref9}
Yuezun Li, Siwei Lyu. 2019.
\textit{Exposing DeepFake Videos By Detecting Face Warping Artifacts}.
\url{https://doi.org/10.48550/arXiv.1811.00656}

\bibitem{ref10}
Sheng-Yu Wang, Oliver Wang, Andrew Owens, Richard Zhang, Alexei A. Efros. 2019.
\textit{Detecting Photoshopped Faces by Scripting Photoshop}.
\url{https://doi.org/10.48550/arXiv.1906.05856}

\bibitem{ref11}
Tackhyun Jung, Sangwon Kim, Keecheon Kim. 2020.
\textit{DeepFake Detection via Facial Landmark Analysis}.
\url{https://doi.org/10.1109/ACCESS.2020.2988660}

\bibitem{ref12}
Umur Aybars Ciftci, Ilke Demir, Lijun Yin. 2020.
\textit{FakeCatcher: Detection of Synthetic Portrait Videos using Biological Signals}.
\url{https://doi.org/10.1109/TPAMI.2020.3009287}

\bibitem{ref13}
Andreas Rössler, Davide Cozzolino, Luisa Verdoliva, Christian Riess, Justus Thies, Matthias Niessner. 2019.
\textit{FaceForensics++: Learning to Detect Manipulated Facial Images}.
\url{https://doi.org/10.1109/ICCV.2019.00009}

\bibitem{ref14}
Darius Afchar, Vincent Nozick, Junichi Yamagishi, Isao Echizen. 2018.
\textit{MesoNet: a Compact Facial Video Forgery Detection Network}.
\url{https://doi.org/10.1109/WIFS.2018.8630761}

\bibitem{ref15}
Shen Chen, Taiping Yao, Yang Chen, Shouhong Ding, Jilin Li, Rongrong Ji. 2021.
\textit{Local Relation Learning for Face Forgery Detection}.
\url{https://doi.org/10.1609/aaai.v35i2.16193}

\bibitem{ref16}
Jiaming Li, Hongtao Xie, Lingyun Yu, Xingyu Gao, Yongdong Zhang. 2023.
\textit{Discriminative Feature Mining Based on Frequency Information and Metric Learning for Face Forgery Detection}.
\url{https://doi.org/10.1109/TKDE.2021.3117003}

\bibitem{ref17}
Luchuan Song, Zheng Fang, Xiaodan Li, Xiaoyi Dong, Zhenchao Jin, Yuefeng Chen, Siwei Lyu. 2022.
\textit{Adaptive Face Forgery Detection in Cross Domain}.
\url{https://doi.org/10.1007/978-3-031-19830-4_27}

\bibitem{ref18}
Tianchen Zhao, Xiang Xu, Mingze Xu, Hui Ding, Yuanjun Xiong, Wei Xia. 2021.
\textit{Learning Self-Consistency for Deepfake Detection}.
\url{https://doi.org/10.1109/ICCV48922.2021.01475}

\bibitem{ref19}
Elahe Vahdani, Yingli Tian. 2023.
\textit{Deep Learning-Based Action Detection in Untrimmed Videos: A Survey}.
\url{https://doi.org/10.1109/TPAMI.2022.3193611}

\bibitem{ref20}
Quan Zhang, Jinwei Fang, Rui Yuan, Xi Tang, Yuxin Qi, Ke Zhang, Chun Yuan. 2025.
\textit{Weakly Supervised Temporal Action Localization via Dual-Prior Collaborative Learning Guided by Multimodal Large Language Models}.
\url{https://doi.org/10.1109/CVPR52734.2025.02248}

\bibitem{ref21}
Mamshad Nayeem Rizve, Gaurav Mittal, Ye Yu, Matthew Hall, Sandra Sajeev, Mubarak Shah, Mei Chen. 2023.
\textit{PivoTAL: Prior-Driven Supervision for Weakly-Supervised Temporal Action Localization}.
\url{https://doi.org/10.1109/CVPR52729.2023.02202}

\bibitem{ref22}
Filipo Sharevski, Rawan Zeidieh. 2024.
\textit{Blind and Low Vision Individuals' Detection of Audio Deepfakes}.
In Proceedings of the 2024 ACM SIGSAC Conference on Computer and Communications Security (CCS '24).
\url{https://doi.org/10.1145/3658644.3670353}

\bibitem{ref23}
Zhan Tong, Yibing Song, Jue Wang, Limin Wang. 2022.
\textit{VideoMAE: Masked Autoencoders are Data-Efficient Learners for Self-Supervised Video Pre-Training}.\url{https://proceedings.neurips.cc/paper_files/paper/2022/file/416f9cb3276121c42eebb86352a4354a-Paper-Conference.pdf}

\bibitem{ref24}
Alexey Dosovitskiy et al. 2021.
\textit{An Image Is Worth 16x16 Words: Transformers for Image Recognition at Scale}.
\url{https://openreview.net/pdf?id=YicbFdNTTy}

\bibitem{ref25}
Shuming Liu, Chen-Lin Zhang, Chen Zhao, Bernard Ghanem. 2024.
\textit{End-to-End Temporal Action Detection with 1B Parameters Across 1000 Frames}.
\url{https://doi.org/10.1109/CVPR52733.2024.01759}

\bibitem{ref26}
Daisuke Niizumi, Daiki Takeuchi, Yasunori Ohishi, Noboru Harada, Kunio Kashino. 2021.
\textit{BYOL for Audio: Self-Supervised Learning for General-Purpose Audio Representation}.
\url{https://doi.org/10.1109/IJCNN52387.2021.9534474}

\bibitem{ref27}
Gao Huang, Zhuang Liu, Laurens Van Der Maaten, Kilian Q. Weinberger. 2017.
\textit{Densely Connected Convolutional Networks}.
\url{https://doi.org/10.1109/CVPR.2017.243}

\bibitem{ref28}
Jun-Tae Lee, Mihir Jain, Hyoungwoo Park, Sungrack Yun. 2021.
\textit{Cross-Attentional Audio-Visual Fusion for Weakly-Supervised Action Localization}.
\url{https://openreview.net/pdf?id=hWr3e3r-oH5}

\bibitem{ref29}
He Wang, Pengcheng Guo, Pan Zhou, Lei Xie. 2024.
\textit{MLCA-AVSR: Multi-Layer Cross Attention Fusion Based Audio-Visual Speech Recognition}.
\url{https://doi.org/10.1109/ICASSP48485.2024.10446769}

\bibitem{ref30}
Komal Chugh, Parul Gupta, Abhinav Dhall, Ramanathan Subramanian. 2020.
\textit{Not Made for Each Other: Audio-Visual Dissonance-based Deepfake Detection and Localization}.
In Proceedings of the 28th ACM International Conference on Multimedia, 439–447.
\url{https://doi.org/10.1145/3394171.3413700}

\bibitem{ref31}
Megha Nawhal, Greg Mori. 2021.
\textit{Activity Graph Transformer for Temporal Action Localization}.
\url{https://doi.org/10.48550/arXiv.2101.08540}

\bibitem{ref32}
Tianwei Lin, Xiao Liu, Xin Li, Errui Ding, Shilei Wen. 2019.
\textit{BMN: Boundary-Matching Network for Temporal Action Proposal Generation}.
\url{https://doi.org/10.1109/ICCV.2019.00399}

\bibitem{ref33}
Anurag Bagchi, Jazib Mahmood, Dolton Fernandes, Ravi Kiran Sarvadevabhatla. 2022.
\textit{Hear Me Out: Fusional Approaches for Audio Augmented Temporal Action Localization}.
\url{https://doi.org/10.48550/arXiv.2106.14118}

\bibitem{ref34}
Jiachen Lu et al. 2021.
\textit{SOFT: Softmax-free Transformer with Linear Complexity}.\url{https://proceedings.neurips.cc/paper_files/paper/2021/file/b1d10e7bafa4421218a51b1e1f1b0ba2-Paper.pdf}

\bibitem{ref35}
Chen-Lin Zhang, Jianxin Wu, Yin Li. 2022.
\textit{ActionFormer: Localizing Moments of Actions with Transformers}.
In European Conference on Computer Vision (ECCV), 492–510.
\url{https://doi.org/10.1007/978-3-031-19772-7_29}

\bibitem{ref36}
Rui Zhang et al. 2023.
\textit{Ummaformer: A Universal Multimodal-Adaptive Transformer Framework for Temporal Forgery Localization}.
\url{https://doi.org/10.1145/3581783.3613767}

\bibitem{ref37}
Christos Koutlis, Symeon Papadopoulos. 2024.
\textit{DiModiF: Discourse Modality-Information Differentiation for Audio-Visual Deepfake Detection and Localization}.
\url{https://doi.org/10.48550/arXiv.2411.10193}

\bibitem{ref38}
Darius Afchar, Vincent Nozick, Junichi Yamagishi, Isao Echizen. 2018.
\textit{Mesonet: a compact facial video forgery detection network}.
In 2018 IEEE International Workshop on Information Forensics and Security (WIFS), pages 1--7. IEEE.
\url{https://doi.org/10.1109/WIFS.2018.8630761}

\bibitem{ref39}
Zhixi Cai, Shreya Ghosh, Abhinav Dhall, Tom Gedeon, Kalin Stefanov, Munawar Hayat. 2023.
\textit{Glitch in the matrix: A large scale benchmark for content driven audio-visual forgery detection and localization}.
Computer Vision and Image Understanding, 236: 103818.
\url{https://doi.org/10.1016/j.cviu.2023.103818}

\bibitem{ref40}
Dingfeng Shi, Yujie Zhong, Qiong Cao, Lin Ma, Jia Li, Dacheng Tao. 2023.
\textit{Tridet: Temporal Action Detection with Relative Boundary Modeling}.
\url{https://doi.org/10.1109/CVPR52729.2023.01808}

\bibitem{ref41}
Y. Zhang et al. 2024.
\textit{MFMS: Learning Modality-Fused and Modality-Specific Features for Deepfake Detection and Localization Tasks}.
\url{https://doi.org/10.1145/3664647.3688984}

\bibitem{ref42}
Ashish Vaswani et al. 2017.
\textit{Attention Is All You Need}.
\url{https://proceedings.neurips.cc/paper_files/paper/2017/file/3f5ee243547dee91fbd053c1c4a845aa-Paper.pdf}

\bibitem{ref43}
Weihao Yu et al. 2024.
\textit{MetaFormer Baselines for Vision}.
IEEE Transactions on Pattern Analysis and Machine Intelligence.
\url{https://doi.org/10.1109/TPAMI.2023.3329173}

\bibitem{ref44}
Zhaohui Zheng, Ping Wang, Wei Liu, Jinze Li, Rongguang Ye, Dongwei Ren. 2020.
\textit{Distance-IoU Loss: Faster and Better Learning for Bounding Box Regression}.
\url{https://doi.org/10.1609/aaai.v34i07.6999}

\bibitem{ref45}
Tsung-Yi Lin, Priya Goyal, Ross Girshick, Kaiming He, Piotr Dollár. 2017.
\textit{Focal Loss for Dense Object Detection}.
\url{https://doi.org/10.1109/ICCV.2017.324}

\bibitem{ref46}
Zonghui Guo, Yingjie Liu, Jie Zhang, Haiyong Zheng, Shiguang Shan. 2025.
\textit{Face Forgery Video Detection via Temporal Forgery Cue Unraveling}.
In 2025 IEEE/CVF Conference on Computer Vision and Pattern Recognition (CVPR).
\url{https://doi.org/10.1109/CVPR52734.2025.00693}

\bibitem{ref47}
Trevine Oorloff, Surya Koppisetti, Nicolò Bonettini, Divyaraj Solanki, Ben Colman, Yaser Yacoob, Ali Shahriyari, Gaurav Bharaj. 2024.
\textit{AVFF: Audio-Visual Feature Fusion for Video Deepfake Detection}.
In 2024 IEEE/CVF Conference on Computer Vision and Pattern Recognition (CVPR).
\url{https://doi.org/10.1109/CVPR52733.2024.02559}

\bibitem{ref48}
Juan Hu, Xin Liao, Difei Gao, Satoshi Tsutsui, Qian Wang, Zheng Qin, Mike Zheng Shou. 2024.
\textit{Delocate: Detection and Localization for Deepfake Videos with Randomly-Located Tampered Traces}.
In Proceedings of the Thirty-Third International Joint Conference on Artificial Intelligence (IJCAI).
\url{https://doi.org/10.24963/ijcai.2024/648}

\bibitem{ref49}
Zijie Lou, Gang Cao, Man Lin, Lifang Yu, Shaowei Weng. 2025.
\textit{Trusted Video Inpainting Localization via Deep Attentive Noise Learning}.
In IEEE Transactions on Dependable and Secure Computing.
\url{https://doi.org/10.1109/TDSC.2025.3595960}

\bibitem{ref50}
Ziyi Liu, Yangcen Liu. 2025.
\textit{Bridge the Gap: From Weak to Full Supervision for Temporal Action Localization with PseudoFormer}.
In 2025 IEEE/CVF Conference on Computer Vision and Pattern Recognition (CVPR).
\url{https://doi.org/10.1109/CVPR52734.2025.00814}

\bibitem{ref51}
Jianyang Xie, Yitian Zhao, Yanda Meng, He Zhao, Anh Nguyen, Yalin Zheng. 2025.
\textit{Are Spatial-Temporal Graph Convolution Networks for Human Action Recognition Over-Parameterized?}.
In 2025 IEEE/CVF Conference on Computer Vision and Pattern Recognition (CVPR).
\url{https://doi.org/10.1109/CVPR52734.2025.02264}

\bibitem{ref52}
Xinfeng Li, Kai Li, Yifan Zheng, Chen Yan, Xiaoyu Ji, Wenyuan Xu. 2024.
\textit{SafeEar: Content Privacy-Preserving Audio Deepfake Detection}.
In Proceedings of the 2024 ACM SIGSAC Conference on Computer and Communications Security (CCS '24).
\url{https://doi.org/10.1145/3658644.3690292}

\bibitem{ref53}
Bowen Zhang, Terence Sim. 2022.
\textit{Localizing Fake Segments in Speech}.
\url{https://doi.org/10.1109/ICPR56361.2022.9956134}

\bibitem{ref54}
Sauradip Nag, Xiatian Zhu, Yi-Zhe Song, Tao Xiang. 2022.
\textit{Proposal-Free Temporal Action Detection via Global Segmentation Mask Learning}.
In European Conference on Computer Vision (ECCV), 645--662.
\url{https://doi.org/10.1007/978-3-031-20062-5_37}

\bibitem{ref55}
Guo Chen, Yin-Dong Zheng, Limin Wang, Tong Lu. 2022.
\textit{DCAN: Improving Temporal Action Detection via Dual Context Aggregation}.
In Proceedings of the AAAI Conference on Artificial Intelligence, 248--257.
\url{https://doi.org/10.1609/aaai.v36i1.19900}

\bibitem{ref56}
Zhixi Cai, Shreya Ghosh, Aman Pankaj Adatia, Munawar Hayat, Abhinav Dhall, Tom Gedeon, Kalin Stefanov. 2024.
\textit{AV-Deepfake1M: A Large-Scale LLM-Driven Audio-Visual Deepfake Dataset}. 
In Proceedings of the 32nd ACM International Conference on Multimedia, 7414--7423.
\url{https://doi.org/10.1145/3664647.3680795}

\end{thebibliography}
\end{document}